# CFPNet-M: A Light-Weight Encoder-Decoder Based Network for Multimodal Biomedical Image Real-Time Segmentation

Ange Lou, Shuyue Guan, and Murray Loew, *Fellow, IEEE*

**Abstract**—**Deep learning techniques are proving instrumental in identifying, classifying, and quantifying patterns in medical images. Segmentation is one of the important applications in medical image analysis. The U-Net has become the predominant deep-learning approach to medical image segmentation tasks. Existing U-Net based models have limitations in several respects, however, including: the requirement for millions of parameters in the U-Net, which consumes considerable computational resources and memory; the lack of global information; and incomplete segmentation in difficult cases. To remove some of those limitations, we built on our previous work and applied two modifications to improve the U-Net model: 1) we designed and added the dilated channel-wise CNN module and 2) we simplified the U-shape network. We then proposed a novel light-weight architecture, the Channel-wise Feature Pyramid Network for Medicine (CFPNet-M). To evaluate our method, we selected five datasets from different imaging modalities: thermography, electron microscopy, endoscopy, dermoscopy, and digital retinal images. We compared its performance with several models having a variety of complexities. We used the Tanimoto similarity instead of the Jaccard index for gray-level image comparisons. The CFPNet-M achieves segmentation results on all five medical datasets that are comparable to existing methods, yet require only 8.8 MB memory, and just 0.65 million parameters, which is about 2% of U-Net. Unlike other deep-learning segmentation methods, this new approach is suitable for real-time application: its inference speed can reach 80 frames per second when implemented on a single RTX 2070Ti GPU with an input image size of 256×192 pixels.**

## I. INTRODUCTION

The goals of medical image analysis are to provide tools for efficient diagnostic and treatment processes for radiologists and clinicians [1]. Medical imaging devices such as X-ray, CT, and MRI can provide anatomic and functional information about diseases and abnormalities inside the body nondestructively. Because many medical images have imperfections, artifacts, and confounding details (and even in the absence of them), it is necessary to process them to extract useful information [2] from them. Image processing technologies have made many contributions in medical applications; for example, image segmentation, image registration, and image-guided surgery are widely used in medical diagnosis and treatment.

In recent years, medical imaging has undergone tremendous developments in high-performance segmentation methods based on deep learning. Early segmentation models are mainly of encoder-decoder based neural networks like FCN [3], UNet [4], and SegNet [5]; they are widely used for segmentation in many scenes. Then, residual connection [6] has been applied to those encoder-decoder based models to help training and to improve the efficiency of networks [7]. To solve the class imbalance problem which is usually caused by imbalanced data and uncertain labels like small organs and vague boundaries [8][53] in segmentation tasks without adding more training parameters, the Inception module [9] and dilated convolution [10] have been introduced to design neural networks. For example, MutliResUNet [11] and DC-UNet [12] have applied the simplified Inception module to the UNet and successfully improved the performance for small-tumor segmentations. To segment objects of different sizes, some networks like LEDNet [13] and ESPNet [14] introduce dilated convolution to extract global feature information.

These new convolutional techniques have improved segmentation performance, but many state-of-art networks like Inception v3 [15], DeepLab [16] and MobileNet v2 [17] have significant computational cost: they contain millions or tens of millions of parameters and thus consume a large amount of computational resources and memory for training and prediction. In the field of semantic segmentation for autonomous driving, there have already been many light-weight networks with very good performance, including ESPNetv2 [18], DABNet [19], and ICNet [20]; the mean accuracy as measured by Intersection over Union (mIoU) of their results can achieve 70% on the Cityscapes [21] dataset. Real-time segmentation has taken on a more important role, with applications in image-guided surgery [22] and medical robotics [23] areas that can benefit from the development of deep learning techniques. For robotic surgery, an effective

Ange Lou is with the Biomedical Engineering Department, George Washington University, Washington, DC 20052 USA (e-mail: angelou@gwmail.gwu.edu).

Shuyue Guan is with the Biomedical Engineering Department, George Washington University, Washington, DC 20052 USA (e-mail: frankshuyueguan@gwu.edu).

Murray Loew is with the Biomedical Engineering Department, George Washington University, Washington, DC 20052 USA (e-mail: loew@gwu.edu).

and real-time image analysis is important for accurate surgical guidance. There are limited studies, however, about real-time segmentation networks in medical image segmentation. In addition, each imaging modality has its specific challenges. Therefore, it is important to build a light-weight real-time segmentation network that is effective across the range of modalities.

In this study, in parallel with improvements of the capabilities of those encoder-decoder based models, we have discovered some techniques to improve segmentation performance and to decrease the number of parameters in models. We developed a novel model called the Channel-wise Feature Pyramid Network for Medicine (CFPNet-M), an extension of our previous work [24]. We tested the performance of our method on datasets from five different modalities and compared the accuracies with results from other networks. Experiment results show that CFPNet-M has competitive accuracy and remarkably fewer parameters.

The contributions of this paper can be summarized as follow:

- We apply dilated convolution and simplify the Inception module to design the CFP module.
- Based on the CFP module, we propose the CFPNet-M, which is a light-weight network for real-time medical image segmentation.
- We experiment with biomedical datasets of different modalities, and CFPNet-M shows competitive accuracy and superiority in terms of the parameter size and prediction speed.
- We examine both region (tumors and polyps)-based and thin (neural structures and vessels) objects to show the ability of CFPNet-M to perform well for complex medical images.
- We introduce the Tanimoto similarity as a measurement metric and compare its performance with other metrics.

## II. RELATED WORK

Recently, many deep learning-based methods have been proposed for image segmentation. Compared with traditional methods, such as texture analysis and shape detection, deep learning-based methods are more automatic, accurate, and adaptive to object segmentation in complicated situations, which are common in medical images. Thus, neural network models are good candidates for semantic segmentation of medical images.

### A. Encoder-Decoder-Based CNNs for Semantic Segmentation

The encoder-decoder-based network can be divided into two parts: encoder and decoder. Usually, the encoder is a sequence of convolutional kernels and down-sampling operators used to extract high-order features. These features are supposed to contain the pixel-wise semantic information about the objects, edges, and background. And then these features are decoded by up-sampling or deconvolutional operators to generate segmentation masks. Usually, the masks show the probabilities of pixels belonging to foreground or background. There are many encoder-decoder-based state-of-the-art implementations including FCN, SegNet and UNet. Especially, the UNet shows great potential in pixel-level medical image segmentation by achieving better performances than others. In addition, the MutliResUNet [11] adds the Residual Paths from ResNet [6] to protect the deeper network (depth increasing, i.e., more hidden layers) from the vanishing-gradient problem during training.

### B. Inception Module

The Inception module [9] proposed a parallel structure that contains 1×1, 3×3, and 5×5 convolution kernels to obtain multi-scale feature maps (Figure 1a). Those large kernels, however, lead to great computational cost. Therefore, the feature version of the Inception module introduced factorization to reduce the number of parameters and the computational cost. The factorization operations contain two parts: factorization into smaller convolution kernels and asymmetric convolutions [15]. These two factorization methods have been applied widely in many networks (e.g., ResNext [25], Xception [26], and MobileNets [27]). The parallel structures in the Inception module increase the width of networks and provide multi-scale features (Figure 1c), and applying the factorization reduces the computational costs of increasing the width. To provide more different-scale (more effective) features for segmentation, we propose the Channel-wise Feature Pyramid (CFP) module, which will be further discussed in the Methods section.

### C. Dilated Convolution

Dilated convolution [28] introduced a special form of standard $3 \times 3$ convolution by inserting gaps between pairs of convolution elements to enlarge the effective receptive field without introducing more parameters. The effective size of an $n \times n$ dilated convolution kernel with a dilation rate $r$ can be represented as: $[r(n-1)+1]^2$. The dilation rate is the number of pixel gaps between adjacent convolution elements, and only $n^2$ parameters participate in model training. In the semantic segmentation area, many studies like ESPNetv2 [18], DABNet [19], and LEDNet [13] have used the dilated convolution kernels in their models and shown its advantage in pixel-level segmentation. The dilated convolution has shown the ability to extract large-scale features while maintaining the total number of parameters [51, 52]. Based on the considerations above, in this study, we apply the dilated convolution in every channel of the CFP module to further decrease parameters and make the model light-weight.

Designing the architectures of models is always a core study in deep learning. Some architectures like the ResNet [6] go deeper to extract higher-level features and some architectures like the InceptionNet [9] go wider to acquire more features at the same level. In general, it is difficult to show, for the deeper and wider networks, which one is the better architecture in deep learning. But for segmentation tasks, we consider choosing a wider architecture because it could provide multi-scale and different features from input images to form the final segmentation masks [15]. Unlike the classification or recognition tasks, image segmentation may not require very high-level features of objects. We think semantic segmentation may require more optional features to decide the boundaries and backgrounds. Thus, we introduce the CFP module for segmentation.

In segmentation applications, most current models are far from running in real-time ($\geq$25 frames per second (FPS)) [45]. For some applications such as autonomous driving, real-time performance is as important as accuracy. And real-time [48-50] and light-weight [46, 47] models are also required by medical segmentation used in some cases, including a real-time surgery-assistant system and small/portable diagnostic devices. Recently, many deep learning segmentation models have been proposed; few, however, balance well the speed, model size, and performance. For example, on the Cityscapes test set, the PSPNet [54] can reach mIoU = 78.4% but the model has 65.7M parameters and its FPS is smaller than 1; in contrast, the ESPNet [14] model only has 0.4M parameters and its FPS is 112 but mIoU = 60.3%. Our CFP module aims to have a better balance among these aspects, especially for application to medical segmentation.

## III.  METHODS

In this section, we firstly introduce the basic element of the CNN module — the Channel-wise Feature Pyramid (CFP) module. Then, we present the structure of the CFP module and architecture of CFPNet-M.

### A.  Feature Pyramid Channel

The CFP module is based on the Feature Pyramid (FP) channel, a factorized form of a convolution operator that decomposes a large kernel into a series of smaller convolution operators as shown in Figures 1(a) and (b). The Inception-v2 applied two $3 \times 3$ convolutional operators to replace the $5 \times 5$ kernel in the naïve version. Based on the idea of multi-scale feature maps, we design a module that contains $3 \times 3$, $5 \times 5$, and $7 \times 7$ kernels as shown in Figure 1(c). Similar to Inception-v2, we applied a series of $3 \times 3$ convolutional operators to replace the $5 \times 5$ and $7 \times 7$ operators. Although this operation saves 28% and 45% of the parameters respectively, there are still too many to achieve the goal of a light-weight model. Then, we applied the technique of MultiRes block [11] and merged those convolution kernels into one channel that contains only three $3 \times 3$ operators. Further, we used the skip connection layer to concatenate features that are extracted from each convolutional operator to build our FP channel, as shown in Figure 1(d). Compared with an Inception-v2-like implementation, the FP channel uses 50% fewer parameters but also retains the similar ability to learn multi-scale feature information.

Because of the concatenating features of each convolutional operator, to keep the same dimension of input and output, we rearranged filter numbers for each operator. For example, if the input dimension is $N$, we assigned $N/4$ for the first and second operators, which correspond to the $3 \times 3$ and $5 \times 5$ convolutions, respectively. For the third block, which is the $7 \times 7$ operator, we assigned $N/2$ filters to extract large-weighted sizeable features.

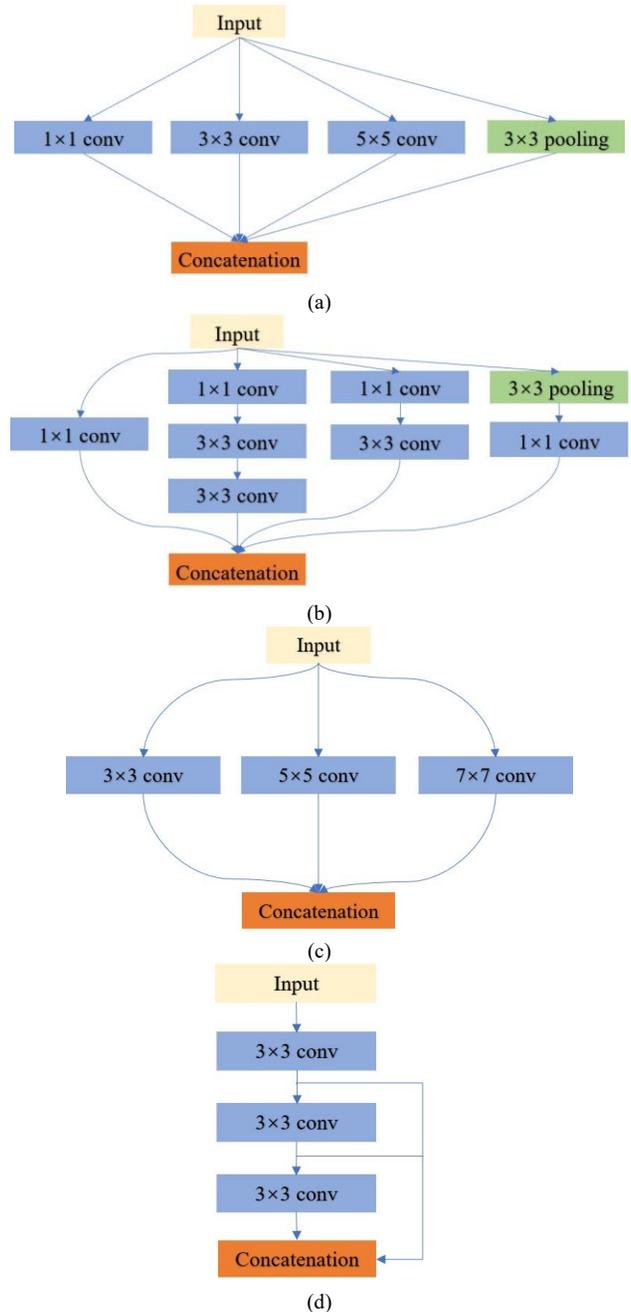

Figure 1. (a) naïve Inception, (b) Inception-v2, (c) Inception-like module, (d) FP Channel,

### B.  CFP Module

The CFP module contains $K$ FP channels with different dilation rates $\{r_1, r_2, ..., r_K\}$. The original CFP module first applied a $1 \times 1$ convolution to reduce the input dimension from $M$ to $M/K$. Then, the dimension of the first to third operators are $M/4K$, $M/4K$, and $M/2K$, respectively.

Figure 2 contains additional details about the CFP module. We use $1 \times 1$ convolution to project high-dimension feature maps to low-dimension. Then we set $K$ FP channels into a parallel structure with different dilation rates. We then concatenate all feature maps into the input's dimension and use another $1 \times 1$ convolution to activate output. This is a basic structure of original CFP module, as shown in Figure 2(a). Factorization increases the depth of network; however, it causes the difficulty of training like gradient vanishing. Moreover, a simple fusion method leads to some unwanted checkerboard or gridding textures that greatly influence the accuracy and quality of segmentation [14]. Thus, we also introduced a hierarchical adding method called Hierarchical Feature Fusion (HFF) [14] that has been shown to eliminate those artifacts. To solve the difficulty of training, we firstly used the residual connection to make a trainable deeper network and then to provide additional feature information [6]. The final version of the CFP module is shown in Figure 2(b).

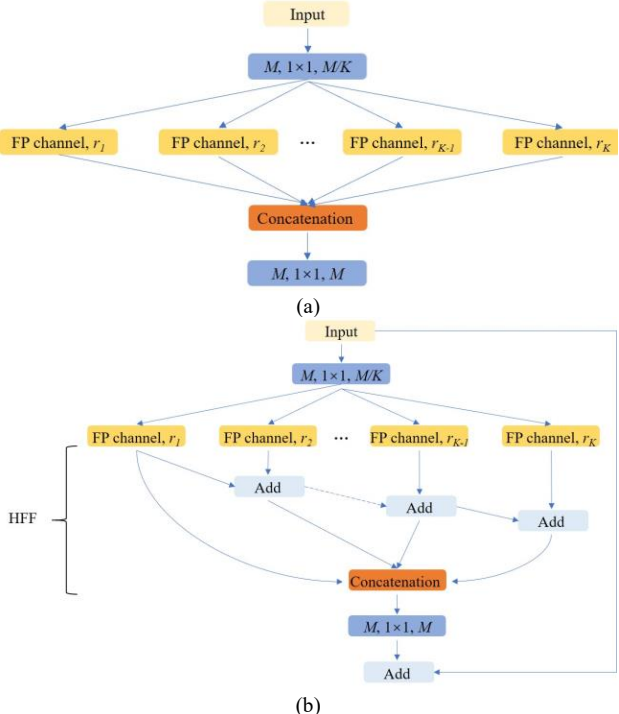

Figure 2. (a) Original CFP module. (b) CFP module

### C. CFPNet-M

**Setup CFP module**: First, we specify the details about the CFP module (Figure 2b) which we used to build the CFPNet-M. We choose the number of FP channels $K = 4$. For the input with dimension $M = 32$, for example, the total number of filters of each channel is 8. And we set the number of filters of the first to third convolutional operators as 2, 2, and 4, respectively. Then, we set different dilation rates for each FP channel. The dilation rate is equal to $r_K$; for example, we set the first and forth channels' dilation rates to $r_1 = 1$ and $r_K$. To extract local and global features, we set the dilation rates of the second and third channels to $r_2 = r_K/4$ and $r_3 = r_K/2$, respectively. Therefore, the CFP module could learn those medium-size features. Note: if $r_K/4 < 1$, e.g., $r_K = 2$, we directly set this channel's dilation rate as 1.

**Network architecture**: Since our goal is to build a light-weight network but have competitive performance, we build a shallow U-shape network as shown in Figure 3. The details of the architecture are in Table I. We firstly use three $3 \times 3$ convolutional operators as the initial feature extractor. The first operators are used with stride 2 to apply down-sampling. After the initial extractor, we applied average pooling for down-sampling. Then, we insert the first CFP module clusters (CFP-M-1), repeating $n$ times. Before the second CFP module clusters (CFP-M-2), we also insert an average-pooling layer to down-sample the feature maps. We repeat the CFP modules $m$ times to build the CFP-M-2 cluster. Before the decoder, we inject resized input images to provide additional feature information for the segmentation network. Then, we apply three deconvolutional operators with stride 2 to build the decoder and connect the same stage encoders by skip connections. Finally, we use a $1 \times 1$ convolution to activate the final feature map and generate segmentation masks. In the CFPNet-M, we choose the repeat times of CFP module $n = 2$ and $m = 6$ with the dilation rate $r_{K_{CFP-M-1}} = [2,2]$ and $r_{K_{CFP-M-2}} = [4,4,8,8,16,16]$.

TABLE I
ARCHITECTURE DETAILS OF CFPNET-M.

| No. | Layer | Mode | Dimension |
|---|---|---|---|
| 1 | $3 \times 3$ Conv | Stride 2 | 32 |
| 2 | $3 \times 3$ Conv | Stride 1 | 32 |
| 3 | $3 \times 3$ Conv | Stride 1 | 32 |
| 4 | Ave pooling | - | - |
| 5-6 | 2×CFP | $r_K = 2$ | 64 |
| 7 | Ave pooling | - | - |
| 8-9 | 2×CFP | $r_K = 4$ | 128 |
| 10-11 | 2×CFP | $r_K = 8$ | 128 |
| 12-13 | 2×CFP | $r_K = 16$ | 128 |
| 14 | Deconv | Stride 2 | 128 |
| 15 | Deconv | Stride 2 | 64 |
| 16 | Deconv | Stride 2 | 32 |
| 17 | $1 \times 1$ Conv | Stride 1 | 1 |

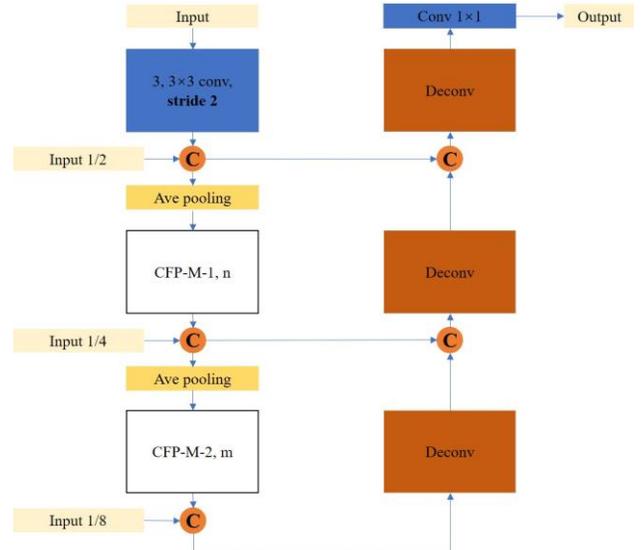

Figure 3. Architecture of CFPNet-M

## IV. EXPERIMENTS

In this section we test our purposed segmentation neural network - CFPNet-M -- on five datasets: our own infrared breast dataset, these datasets that are used widely to evaluate medical segmentation methods: the ISBI-2012 (Electron

Microscopy), ISIC-2018 (Dermoscopy), CVC-ClinicDB (Endoscopy), and DRIVE (Digital Retinal Images).

## A. Datasets

**Our Breast Thermography Images.** We collected breast thermal (8-12 nm) infrared images by using the N2 Imager (N2 Imaging System, Irvine, Calif.). Our breast dataset contains 450 infrared images from 14 patients and 16 healthy volunteers; all images have had the background removed and been denoised. Each participant was imaged for a total time of 15 minutes, capturing one image every minute (15 images per participant). The original sizes of images range from $540 \times 260$ to $610 \times 290$ (pixels); we have uniformly resized them to $256 \times 128$ before training.

**Electron Microscopy Images.** To show the performance of CFPNet-M on electron microscopy (EM) images, we used the dataset of the ISBI-2012: 2D EM segmentation challenge [29, 30]. This dataset contains 30 images in the training set from a serial section Transmission Electron Microscopy (ssTEM) of the Drosophila first instar larva ventral nerve cord [30]. We chose those 30 images for the dataset and resized resolution from $512 \times 512$ to $256 \times 256$.

**Endoscopy Images.** We used CVC-ClinicDB [31] to show the performance of CFPNet-M applied to endoscopy images. Those images were extracted from the colonoscopy videos. The dataset contains a total of 612 images with original size $384 \times 288$. We resized them to $256 \times 192$ for training.

**Dermoscopy Images.** We acquired the Dermoscopy images from the ISIC-2018: Lesion Boundary Segmentation challenge dataset. The data for ISIC-2018 were from the ISIC-2017 [32] and the HAM10000 dataset [33]. The compiled dataset contains a total of 2594 images of different types of skin lesions with expert annotations. Before training, we resized those images to $256 \times 192$.

**Digital Retinal Images.** We applied DRIVE [34] to test the performance of CFPNet-M on thin objects. This dataset was established to help study cardiovascular and ophthalmologic diseases such as diabetes, hypertension, and arteriosclerosis. We used 20 RGB images from its training set and resized them from $512 \times 512$ to $256 \times 256$.

The summarized details of these five datasets can be found in Table II. (Note: All datasets are resized to be smaller according to its original ratio)

## B. Cross-validation

Cross-validation is used widely to test models' performance. In the k-fold cross-validation test, the dataset $D$ is randomly split into $k$ mutually exclusive subsets $D_1, D_2, \dots, D_k$ of approximately equal size [35]. The model is tested k times; for each time, one of the $k$ subsets is chosen as the validation set and other $k - 1$ subsets as the training set. We estimated the performance of model via overall results of $k$-times training. For the infrared breast dataset, we select $k = 30$ because there are 30 individual participants. And we choose $k = 5$ for the other four public datasets.

## C. Measurement metric

To evaluate the performance of segmentation, we need a method to compare the segmented region with ground truth. Since we applied the sigmoid function to activate the final convolutional operator, the output is a gray-level image which maps into the range [0,1]. Therefore, we must threshold before calculating accuracy. Usually, thresholding grayscale image to binary (binarization) [36] introduces additional errors. There are many popular methods to compare two images:

- Binary to binary: Jaccard Similarity (JS) [37]
- Gray to gray: Mean Absolute Error (MAE) [38]; Tanimoto Similarity [39] (extended Jaccard Similarity); Structural Similarity (SSIM) [40].

In our previous studies, we chose JS as the measurement metric; it compares two binary images as two sets A and B, their JS value is:

$$JS(A,B) = \frac{|A \cap B|}{|A \cup B|} \qquad (1)$$

Before comparison, we did threshold using Otsu [41] algorithm. The workflow is shown in the Figure 4.

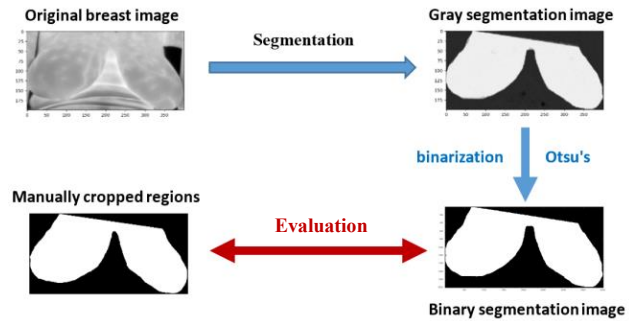

Figure 4. Workflow of Jaccard similarity.

Since the Otsu method led to some thresholding errors, we then considered some gray-to-gray comparison methods. One of simply and widely used method is the Mean Absolute Error (MAE). In general, for two images (same size) $A$ and $B$, their MAE value is:

$$MAE(A,B) = \frac{|A - B|}{maxE} \qquad (2)$$

The **$maxE$** is maximum error value, for 8-bit gray-scale images (size: $W \times L$), $maxE = W \times L \times 2^8$.

TABLE II
OVERVIEW OF DATASETS

| Modality | Dataset | No. of images | Original resolution | Input resolution |
|---|---|---|---|---|
| Thermography | IR breast | 450 | Variable | $256 \times 128$ |
| Electron microscopy | ISBI-2012 | 30 | $512 \times 512$ | $256 \times 256$ |
| Endoscopy | CVC-Clinic DB | 612 | $384 \times 288$ | $256 \times 192$ |
| Dermoscopy | ISCI-2018 | 2594 | Variable | $256 \times 192$ |
| Digital Retinal | DRIVE | 20 | $512 \times 512$ | $256 \times 256$ |

The Tanimoto similarity, also called extended JS, can be considered as a grayscale JS. For binary images, JS compares images by union and intersection operations. The intersection operation could be considered as sum of products. For two set **A** and **B**:

$$|A \cap B| = \sum a_i b_i \qquad (3)$$

Where $a_i \in A, b_i \in B$. This equation holds if $a_i, b_i \in \{0, 1\}$, which are binary values. But if $a_i, b_i$ are not binary, we use sum of products (right part) instead of the union operation. Since:

$$|A \cap A| = \sum a_i^2 \qquad (4)$$

And,

$$|A \cup B| = |A| + |B| - |A \cap B| = \sum (a_i^2 + b_i^2 - a_i b_i) \quad (5)$$

For gray-gray comparison, according to JS(A,B), the value of the Tanimoto similarity is:

$$T(A, B) = \frac{\sum a_i b_i}{\sum (a_i^2 + b_i^2 - a_i b_i)} \qquad (6)$$

To prove that the Tanimoto similarity is an ideal measurement metric, we tested different measurement metrics on different sizes and object-area ratio images as shown in Figure 5(a). Each result is the average value of all (15) samples of one case from our infrared breast dataset. For each sample, the value is calculated by comparing the ground truth image with C-DCNN segmented image [41]. From Figure 5(b) and (c), the results show that SSIM is not stable to changing image size and only the Tanimoto similarity is stable with changes of object-area ratio.

Furthermore, Figure 5(d) shows comparison results of Tanimoto similarity, JS, and MAE for the 15 samples of size 400×200. Results indicate that for a majority of the samples (9/15, noted in yellow), the Tanimoto similarity values are close to JS. Therefore, Tanimoto similarity is a good alternative measure instead of JS for grayscale image comparisons. Based on our analysis, we chose Tanimoto similarity instead of Jaccard as the measurement metric in our experiments.

### D. Implementation protocol

CFPNet-M is trained using Keras [42] with CUDA 9.0, cuDNN V7, and a single RTX 2070Ti GPU. For network training, we chose Adam as the optimizer with parameter $\beta_1 = 0.9$ and $\beta_2 = 0.999$. We set the initial learning rate to 0.001 and used the binary cross-entropy as the loss function.

## V. RESULTS

In the experiments, we chose 10 neural networks with the number of parameters sorted into one of three levels (less than one million, millions, and tens of millions). The details of those networks are shown in Table 3. We tested the performance of U-Net [4], Inception v3 [15], MultiResUNet [11], EfficientNet_b0 [43], DC-UNet [12], MobileNet v2 [17], ICNet[20], ESPNet [14], ENet [44], and our CFPNet-M on five datasets; and then compared their results.

### A. Segmentation results

The IR breast dataset contains images from 30 persons; we divided them into 30 subsets (one subset for one person) for 30-fold cross-validation. The Tanimoto accuracy of each subset is shown in Table VIII in the Appendix. We report the mean Tanimoto similarity and standard deviation (Std) of those networks in Table III. The best mean Tanimoto is shown in bold font.

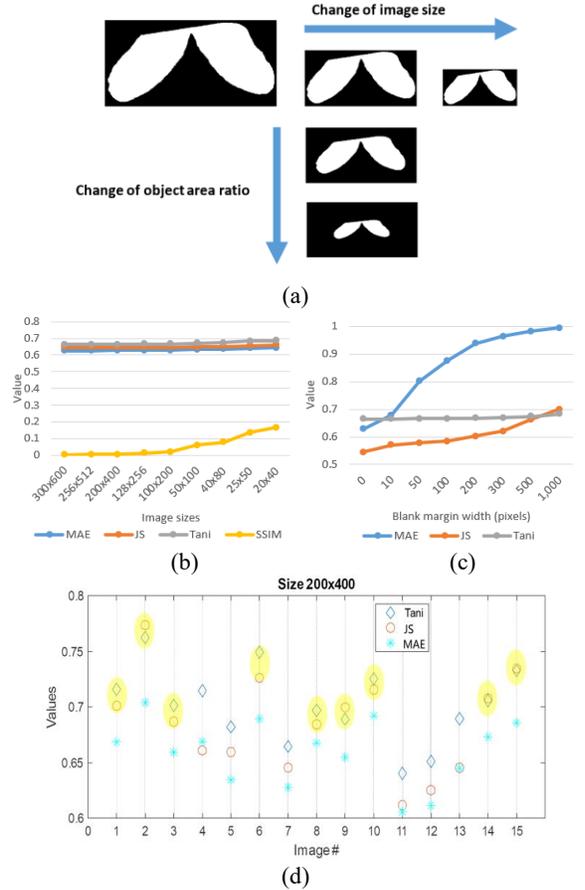

Figure 5. (a) size and ratio change of images. (b) accuracy value vs image size. (c) accuracy vs object ratio. (d) comparison of three measurement metrics

For the four public medical datasets, we divided them randomly into 5 folds and calculated mean Tanimoto similarity and Std. The results are shown in Tables IV to VII. In all experiments, we did not apply additional data augmentation and enhancement techniques. And for the DRIVE dataset, we do not use any post-processing (e.g., extracting objects from field of view). We also display some segmentation results of thin objects in the Figure 6 and Figure 7. And the segmentation results of region-based objects are shown in Figure 8 to 10.

### B. Inference speeds and complexity comparison

Inference speed evaluation was performed on a single Nvidia RTX 2070Ti GPU, we report an average of 500 frames of CVC-ClinicDB dataset for the Frames Per Second (FPS) measurement. To compare the complexity of models, we calculated the floating-point operations (FLOPs) of each network. We report the speeds and complexity of each network in Table IX.

Table VII
RESULTS OF DIGITAL RETINAL DATASETS (DRIVE)

| Model | Fold 1 | Fold 2 | Fold 3 | Fold 4 | Fold 5 | Mean Tanimoto (%) | Std |
|-------|--------|--------|--------|--------|--------|-------------------|-----|
| U-Net | 55.35 | 46.80 | 57.87 | 58.66 | 52.05 | 54.15 | 0.0434 |
| Inception v3 | 52.23 | 57.51 | 57.92 | 56.95 | 56.38 | 56.20 | 0.0205 |
| MultiResUNet | 59.31 | 58.41 | 59.40 | 60.12 | 56.28 | 58.70 | 0.0133 |
| EfficientNet-b0 | 58.57 | 57.40 | 58.97 | 58.17 | 54.12 | 57.44 | 0.0174 |
| DC-UNet | 60.89 | 59.59 | 60.54 | 60.16 | 57.93 | **59.82** | 0.0104 |
| MobileNet v2 | 53.87 | 57.86 | 59.56 | 56.74 | 57.72 | 57.15 | 0.0187 |
| ICNet | 44.22 | 47.63 | 40.16 | 45.97 | 41.67 | 43.93 | 0.0273 |
| ESPNet | 54.07 | 57.30 | 55.25 | 57.32 | 49.75 | 54.74 | 0.0279 |
| ENet | 55.97 | 56.95 | 53.22 | 56.46 | 52.00 | 54.92 | 0.0195 |
| CFPNet-M | 58.59 | 58.17 | 58.32 | 59.37 | 57.22 | 57.15 | 0.0069 |

Table IX
SPEED AND COMPLEXITY OF EACH MODEL

| Model | Parameters | Model file size (MB) | Speed (FPS) | FLOPS |
|-------|-----------|---------------------|-------------|-------|
| U-Net | 31,031,685 | 354.0 | 50 | 464,925,202 |
| Inception v3 | 29,896,689 | 343.0 | 33 | 358,813,094 |
| MultiResUNet | 29,061,741 | 333.0 | 21 | 108,715,188 |
| EfficientNet-b0 | 10,071,501 | 116.0 | 54 | 120,862,094 |
| DC-UNet | 10,069,640 | 116.0 | 17 | 160,517,331 |
| MobileNet v2 | 8,011,345 | 92.5 | 81 | 96,154,791 |
| ICNet | 6,740,481 | 77.7 | 130 | 565,733,942 |
| ESPNet | 579,961 | 7.3 | 129 | 8,668,267 |
| ENet | 365,638 | 5.1 | 168 | 5,370,508 |
| CFPNet-M | 654,279 | 8.8 | 80 | 9,740,456 |

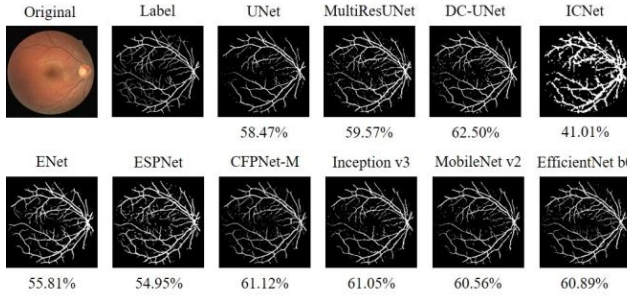

Figure 6. Segmentation of DRIVE.

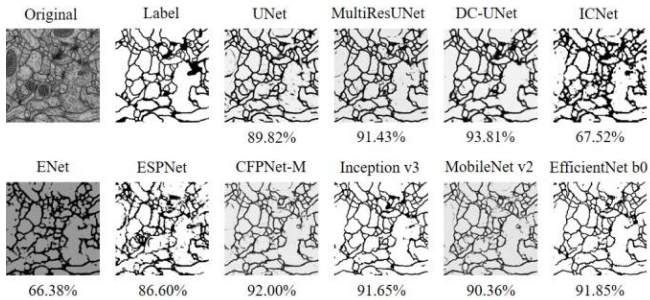

Figure 7. Segmentation of ISBI-2012.

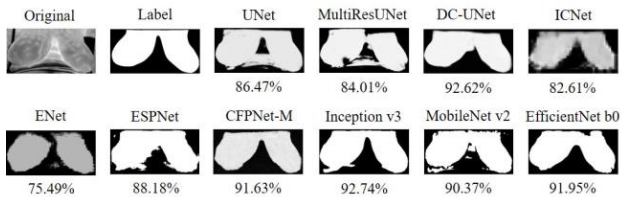

Figure 8. Segmentation of Infrared breast dataset.

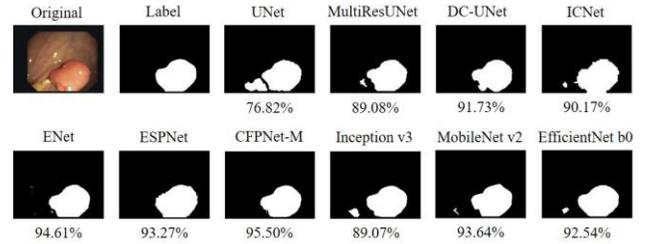

Figure 9. Segmentation of CVC-ClinicDB.

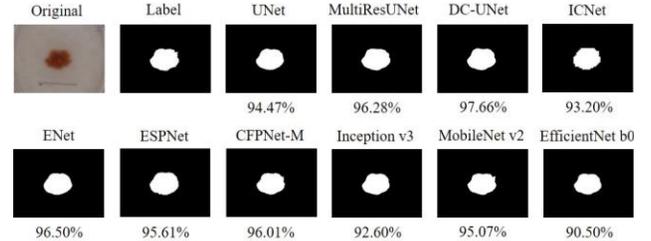

Figure 10. Segmentation of ISIC-2018.

## VI. DISCUSSION

In these experiments, CFPNet-M shows great potential in both region based and thin object medical image segmentation among models whose numbers of parameters are fewer than 1 million (light-weight models). In our infrared breast dataset, as shown in Figure 8, only CFPNet-M can well separate the belly and breast region in light-weight models. According to the average accuracy of infrared breast dataset, the segmentation performance of CFPNet-M is competitive with those state-of-the-arts models in semantic segmentation, like the Inception v3. It is noteworthy that the number of parameters of CFPNet-M is only about 2.2%, 6.5% and, 8.2% of that of Inception v3, EfficientNet-b0, and MobileNet v2 models.

TABLE III
DETAILS OF NETWORKS IN EXPERIMENTS

| Model | Parameters | Model file size (MB) | Mean Tanimoto (%) | Std |
|-------|-----------|---------------------|-------------------|-----|
| U-Net* | 7,750,821 | 88.8 | 88.70 | 0.0256 |
| U-Net | 31,031,685 | 354.0 | 89.80 | 0.0241 |
| Inception v3 | 29,896,689 | 343.0 | **93.17** | 0.0279 |
| MultiResUNet | 29,061,741 | 333.0 | 91.50 | 0.0237 |
| EfficientNet-b0 | 10,071,501 | 116.0 | 92.70 | 0.0272 |
| DC-UNet | 10,069,640 | 116.0 | 92.70 | 0.0215 |
| MobileNet v2 | 8,011,345 | 92.5 | 92.25 | 0.0278 |
| ICNet | 6,740,481 | 77.7 | 84.80 | 0.0458 |
| ESPNet | 579,961 | 7.3 | 89.70 | 0.0250 |
| ENet | 365,638 | 5.1 | 82.90 | 0.0479 |
| CFPNet-M | 654,279 | 8.8 | 92.20 | 0.0211 |

Note: U-Net* is the original U-Net architecture, the U-Net we used in experiments has doubled filter numbers [12]. The model file size is measured with fixed input size $256 \times 128$.

TABLE IV
RESULTS OF EM DATASETS (ISBI-2012)

| Model | Fold 1 | Fold 2 | Fold 3 | Fold 4 | Fold 5 | Mean Tanimoto (%) | Std |
|-------|--------|--------|--------|--------|--------|-------------------|-----|
| U-Net | 90.75 | 89.31 | 91.33 | 92.65 | 91.60 | 91.13 | 0.0157 |
| Inception v3 | 89.51 | 88.51 | 89.40 | 88.16 | 87.56 | 88.63 | 0.0074 |
| MultiResUNet | 90.83 | 91.16 | 92.64 | 93.11 | 92.08 | 91.96 | 0.0121 |
| EfficientNet-b0 | 88.63 | 89.07 | 87.55 | 88.73 | 88.43 | 88.48 | 0.0051 |
| DC-UNet | 91.79 | 91.27 | 93.21 | 93.44 | 93.38 | **92.62** | 0.0092 |
| MobileNet v2 | 89.06 | 88.97 | 88.93 | 89.23 | 89.12 | 89.06 | 0.0011 |
| ICNet | 69.00 | 66.51 | 71.20 | 71.92 | 69.71 | 69.67 | 0.0189 |
| ESPNet | 80.64 | 82.55 | 86.31 | 85.91 | 86.04 | 84.29 | 0.0179 |
| ENet | 64.02 | 72.63 | 66.14 | 64.78 | 69.13 | 67.34 | 0.0276 |
| CFPNet-M | 90.42 | 90.42 | 92.57 | 91.39 | 92.42 | 91.44 | 0.0111 |

TABLE V
RESULTS OF ENDOSCOPY DATASETS (CVC-ClinicDB)

| Model | Fold 1 | Fold 2 | Fold 3 | Fold 4 | Fold 5 | Mean Tanimoto (%) | Std |
|-------|--------|--------|--------|--------|--------|-------------------|-----|
| U-Net | 74.03 | 70.81 | 67.96 | 63.26 | 71.52 | 69.52 | 0.0368 |
| Inception v3 | 86.28 | 86.95 | 83.26 | 81.51 | 83.43 | 84.29 | 0.0203 |
| MultiResUNet | 81.82 | 80.34 | 79.57 | 74.23 | 78.66 | 78.92 | 0.0257 |
| EfficientNet-b0 | 87.32 | 87.31 | 85.72 | 84.39 | 85.64 | **86.08** | 0.0112 |
| DC-UNet | 83.11 | 82.51 | 81.10 | 78.60 | 80.19 | 80.94 | 0.0162 |
| MobileNet v2 | 84.44 | 85.50 | 83.02 | 81.03 | 82.28 | 83.25 | 0.0158 |
| ICNet | 71.00 | 72.60 | 69.52 | 61.28 | 68.60 | 68.60 | 0.0390 |
| ESPNet | 55.55 | 59.57 | 56.30 | 54.77 | 58.48 | 56.94 | 0.0181 |
| ENet | 77.14 | 73.17 | 63.58 | 66.10 | 70.35 | 70.01 | 0.0485 |
| CFPNet-M | 80.31 | 81.86 | 78.09 | 75.43 | 78.49 | 78.84 | 0.0217 |

Table VI
RESULTS OF DERMOSCOPY DATASETS (ISIC-2018)

| Model | Fold 1 | Fold 2 | Fold 3 | Fold 4 | Fold 5 | Mean Tanimoto (%) | Std |
|-------|--------|--------|--------|--------|--------|-------------------|-----|
| U-Net | 79.68 | 79.93 | 80.40 | 80.64 | 76.73 | 79.48 | 0.0141 |
| Inception v3 | 82.36 | 82.32 | 82.63 | 83.54 | 82.74 | 82.70 | 0.0041 |
| MultiResUNet | 80.75 | 80.23 | 80.51 | 81.54 | 80.12 | 80.63 | 0.0051 |
| EfficientNet-b0 | 83.06 | 83.28 | 83.62 | 83.94 | 83.13 | **83.41** | 0.0033 |
| DC-UNet | 80.21 | 81.80 | 82.62 | 82.99 | 80.63 | 81.65 | 0.0108 |
| MobileNet v2 | 81.38 | 82.21 | 82.19 | 82.70 | 81.92 | 82.08 | 0.0043 |
| ICNet | 78.02 | 78.48 | 76.89 | 80.80 | 78.08 | 78.45 | 00129 |
| ESPNet | 76.49 | 78.30 | 73.41 | 80.84 | 75.45 | 76.90 | 0.0253 |
| ENet | 79.01 | 80.29 | 79.95 | 79.78 | 79.77 | 79.76 | 0.0042 |
| CFPNet-M | 81.07 | 81.68 | 81.75 | 83.06 | 81.83 | 81.88 | 0.0065 |

To thin objects datasets like ISBI-2012 (Table IV) and DRIVE (Table VII), the DC-UNet [12] model that proposed in our previous studies performs best. In fact, the CFPNet-M derives from the DC-UNet with much fewer parameters. The CFPNet-M has only about 6.5% parameters of the DC-UNet, but its mean accuracy is slightly smaller than the DC-UNet; and CFPNet-M also shows a good segmentation performance compared with other models we tested.

For ISIC-2018 dataset (region-based segmentation, Table VI), CFPNet-M shows competitive segmentation accuracy but using much fewer parameters. For example, CFPNet-M obtains the mean accuracy very close to the MobileNet v2 and from Table IX, its speed is also similar to the MobileNet v2. However, CFPNet-M has only about 8.2% parameters of the MobileNet v2. Another example is, on the CVC-ClinicDB dataset (Table V), which contains many challenge cases with small objects, vague boundaries, and unbalanced illumination, although the Inception v3, MobileNet v2, and EfficientNet-b0 have better segmentation results than others, CFPNet-M still obtains competitive segmentation results on those non-challenge cases as shown in Figure 9.

The remarkable advantage of CFPNet-M is it has much fewer parameters than those state-of-the-arts models but retains competitively good segmentation performances. It makes the CFPNet-M a light-weight network and the real-time segmentation for multimodal biomedical images to become realizable. We used split-and-merge strategy to reduce the dimension of input and output to minimize the parameters of network. Moreover, we introduced the Hierarchical Feature Fusion (HFF) to eliminate the gridding textures to further improve the segmentation quality. Nevertheless, the multi-channel design enlarges the receptive field and builds a more effective feature pyramid for the encoder. To take a CFP-M module with the dilated rate = 16 as an example, the receptive field at the fourth channel can reach $103 \times 103$ pixels, which is larger than $33 \times 33$ in the ESPNet.

In addition, we report the size, speed, and complexity of each model in Table IX. CFPNet-M not only has fewer parameters and smaller network size but also has lower complexity and higher speeds compared with many state-of-the-art models. Competitive performance, high speed, small size, and low network complexity makes CFPNet-M more suitable to be deployed on portable devices.

## VII. CONCLUSION

In this paper, we analyzed the architecture of the proposed CFPNet-M and compared its performance with some state-of-the-art networks. We selected four public datasets and our own infrared breast dataset to test and compare the performances of ten models including the CFPNet-M for medical image segmentation. The CFPNet-M shows competitive performances and has great advantages of much fewer parameters and smaller model file size. In addition, we also tested its segmentation speeds -- it can reach 80 FPS with $256 \times 192$ input size on a single RTX 2070Ti GPU. Its great real-time performance could help medical applications such as the image-guided surgery and vision based medical robots. Therefore, we believe that the CFPNet-M is a novel and promising light-weight network for real-time medical image segmentations.

# APPENDIX

Table VIII
AVERAGE SEGMENTATION ACCURACY FOR EACH SAMPLE

| | P1 | P2 | P3 | P4 | P5 | P6 | P7 | P8 | P9 | P10 | P11 | P12 | P13 | P14 | V1 | V16 |
|---|---|---|---|---|---|---|---|---|---|---|---|---|---|---|---|---|
| U-Net* | 84.5 | 86.4 | 86.8 | 92.7 | 88.2 | 89.8 | 88.9 | 79.1 | 87.6 | 93.5 | 84.8 | 89.5 | 92.5 | 86.1 | 92.3 | 87.2 |
| U-Net | 85.2 | 87.9 | 87.0 | 94.4 | 90.7 | 91.4 | 84.3 | 79.8 | 89.2 | 94.0 | 87.0 | 93.7 | 93.5 | 87.0 | 93.1 | 87.9 |
| Inception v3 | 89.8 | 93.5 | 88.7 | 95.2 | 91.1 | 91.4 | 90.4 | 84.3 | 94.3 | 96.1 | 93.2 | 95.7 | 90.4 | 93.0 | 95.5 | 95.2 |
| MultiResUNet | 92.9 | 92.6 | 90.0 | 96.7 | 92.1 | 93.6 | 90.4 | 80.0 | 94.8 | 93.0 | 92.7 | 95.6 | 95.0 | 87.6 | 95.5 | 89.0 |
| EfficientNet b0 | 90.3 | 94.7 | 91.1 | 96.7 | 90.7 | 93.0 | 90.0 | 83.4 | 91.4 | 95.5 | 92.5 | 96.2 | 95.9 | 87.6 | 95.7 | 91.0 |
| DC-UNet | 90.3 | 93.2 | 89.0 | 94.8 | 92.4 | 92.4 | 89.0 | 83.3 | 93.8 | 94.9 | 90.8 | 93.8 | 94.2 | 85.3 | 94.8 | 94.2 |
| MobileNet v2 | 86.7 | 83.7 | 84.3 | 85.3 | 86.4 | 87.2 | 81.7 | 87.8 | 87.8 | 88.8 | 80.1 | 88.4 | 84.3 | 87.3 | 86.9 | 83.1 |
| ICNet | 86.7 | 83.7 | 84.0 | 85.3 | 85.7 | 85.3 | 81.7 | 83.3 | 89.8 | 88.8 | 82.0 | 85.6 | 82.6 | 85.3 | 85.1 | 85.1 |
| ESPNet | 87.0 | 85.7 | 88.5 | 89.7 | 85.0 | 85.3 | 88.5 | 78.3 | 86.6 | 73.2 | 73.2 | 83.5 | 82.7 | 88.9 | 87.7 | 93.6 |
| CFPNet-M | 87.5 | 89.7 | 93.0 | 93.8 | 90.4 | 92.0 | 90.2 | 88.2 | 91.7 | 93.9 | 89.5 | 95.5 | 93.6 | 93.4 | 95.0 | 91.2 |

| | V2 | V3 | V4 | V5 | V6 | V7 | V8 | V9 | V10 | V11 | V12 | V13 | V14 | V15 | V16 |
|---|---|---|---|---|---|---|---|---|---|---|---|---|---|---|---|
| U-Net* | 89.5 | 91.9 | 91.0 | 88.7 | 84.1 | 91.7 | 82.6 | 95.1 | 82.5 | 94.1 | 93.3 | 90.4 | 92.3 | 92.3 | 87.2 |
| U-Net | 92.2 | 91.9 | 91.9 | 88.7 | 91.5 | 92.5 | 84.3 | 95.7 | 82.7 | 94.7 | 94.8 | 91.8 | 93.5 | 87.9 | 87.9 |
| Inception v3 | 95.4 | 90.8 | 94.5 | 93.6 | 90.1 | 94.3 | 87.6 | 96.0 | 97.3 | 97.3 | 96.5 | 93.2 | 93.2 | 95.2 | 95.2 |
| MultiResUNet | 94.8 | 92.7 | 92.5 | 90.3 | 93.9 | 93.9 | 87.6 | 96.0 | 88.4 | 96.5 | 95.7 | 93.0 | 93.0 | 89.0 | 89.0 |
| EfficientNet b0 | 94.8 | 94.2 | 94.5 | 93.0 | 91.6 | 95.5 | 88.0 | 95.9 | 91.1 | 95.0 | 95.6 | 94.1 | 94.2 | 91.0 | 91.0 |
| DC-UNet | 93.6 | 91.7 | 92.0 | 91.7 | 92.8 | 94.2 | 89.0 | 95.0 | 89.0 | 96.5 | 93.1 | 91.8 | 85.6 | 94.8 | 94.8 |
| MobileNet v2 | 83.7 | 81.9 | 83.8 | 85.7 | 87.7 | 87.0 | 87.0 | 89.8 | 89.8 | 88.9 | 85.6 | 84.3 | 84.7 | 85.1 | 83.1 |
| ICNet | 85.7 | 82.4 | 82.8 | 81.2 | 81.7 | 81.7 | 82.5 | 84.5 | 86.7 | 82.2 | 82.2 | 82.6 | 82.6 | 85.1 | 85.1 |
| ESPNet | 89.7 | 89.7 | 89.7 | 89.7 | 85.3 | 84.9 | 87.0 | 94.5 | 96.7 | 82.2 | 82.2 | 82.9 | 91.8 | 87.7 | 93.6 |
| CFPNet-M | 89.4 | 91.6 | 90.4 | 93.8 | 92.0 | 88.9 | 88.9 | 90.8 | 95.2 | 96.1 | 91.8 | 92.4 | 92.4 | 95.8 | 91.2 |

Bold value is the maximum for each participant. "P" is patient and "V" is volunteer.